\documentclass{article}

\usepackage{microtype}
\usepackage{graphicx}
\usepackage{subfigure}
\usepackage{booktabs} %

\usepackage{hyperref}

\usepackage[accepted]{icml2025}
\usepackage{xurl}

\usepackage{amsmath}
\usepackage{amssymb}
\usepackage{mathtools}
\usepackage{amsthm}

\usepackage[capitalize,noabbrev]{cleveref}

\usepackage{cuted}
\usepackage{capt-of}
\usepackage{pgfplots}
\usepackage{colortbl}
\usepackage{graphicx}
\usepackage{subcaption}
\usepackage{enumitem}
\usepackage{multicol}
\usepackage{multirow}
\usepackage{float}

\usepackage[most]{tcolorbox}
\usepackage{xltabular}

\usepackage[colorinlistoftodos,prependcaption,textsize=tiny]{todonotes}

\usepackage{framed}
\usepackage[framemethod=TikZ]{mdframed}
\usepackage{wrapfig}
\usepackage{microtype}

\definecolor{blue}{HTML}{BDD8FF}

\theoremstyle{plain}

\theoremstyle{definition}

\theoremstyle{remark}

\usepackage[textsize=tiny]{todonotes}

\icmltitlerunning{WeGeFT (wee-gift): Weight-Generative Fine-Tuning}

\begin{document}

\twocolumn[
\icmltitle{WeGeFT: Weight‑Generative Fine‑Tuning for \\ Multi‑Faceted Efficient Adaptation of Large Models}

\icmlsetsymbol{equal}{*}

\begin{icmlauthorlist}
\icmlauthor{Chinmay Savadikar}{yyy}
\icmlauthor{Xi Song}{xxx}
\icmlauthor{Tianfu Wu}{yyy}\\
\icmlauthor{{\tt Code: \url{https://savadikarc.github.io/wegeft}}}{}
\end{icmlauthorlist}

\icmlaffiliation{yyy}{Department of Electrical and Computer Engineering, North Carolina State University, Raleigh, USA}
\icmlaffiliation{xxx}{An Independent Researcher}

\icmlcorrespondingauthor{Chinmay Savadikar}{csavadi@ncsu.edu}
\icmlcorrespondingauthor{Tianfu Wu}{twu19@ncsu.edu}

\icmlkeywords{Machine Learning, ICML}

\vskip 0.3in
]

\printAffiliationsAndNotice{}  %

\begin{abstract}
Fine-tuning large pretrained Transformer models can focus on either introducing a small number of new learnable parameters (parameter efficiency) or editing representations of a small number of tokens using lightweight modules (representation efficiency). While the pioneering method LoRA (Low-Rank Adaptation) inherently balances parameter, compute, and memory efficiency, many subsequent variants trade off compute and memory efficiency and/or performance to further reduce fine-tuning parameters.
To address this limitation and unify parameter-efficient and representation-efficient fine-tuning, we propose Weight-Generative Fine-Tuning (WeGeFT, pronounced  \textit{wee-gift}), a novel approach that \textbf{learns to generate fine-tuning weights directly from the pretrained weights}. WeGeFT employs a simple low-rank formulation consisting of two linear layers, either shared across multiple layers of the pretrained model or individually learned for different layers. This design achieves multi-faceted efficiency in parameters, representations, compute, and memory, while maintaining or exceeding the performance of LoRA and its variants. Extensive experiments on commonsense reasoning, arithmetic reasoning, instruction following, code generation, and visual recognition verify the effectiveness of our proposed WeGeFT. %
\end{abstract}

\setlength{\abovecaptionskip}{0pt}
\setlength{\belowcaptionskip}{-4pt}
\setlength{\belowdisplayskip}{2pt} \setlength{\belowdisplayshortskip}{2pt}
\setlength{\abovedisplayskip}{2pt} \setlength{\abovedisplayshortskip}{2pt}

\section{Introduction}
\label{sec:intro}

Fine-tuning pretrained deep neural networks (DNNs) as feature backbones for downstream tasks has been an important and challenging research topic. In recent years, large feature backbones with open weights, such as LLaMA~\citep{llama,llama2,llama3}, have become ubiquitous. Training such models from scratch is infeasible with limited resources, and fine-tuning them entirely can also be prohibitively costly.
This raises two critical questions: (i) which parts of a pretrained model should be fine-tuned (often treated as a hyperparameter), and (ii) how those parts should be fine-tuned. In this paper, we focus on the latter question by leveraging module/layer selection strategies widely adopted in prior art.

\begin{figure*} [t]
    \centering
    \includegraphics[width=0.9\linewidth]{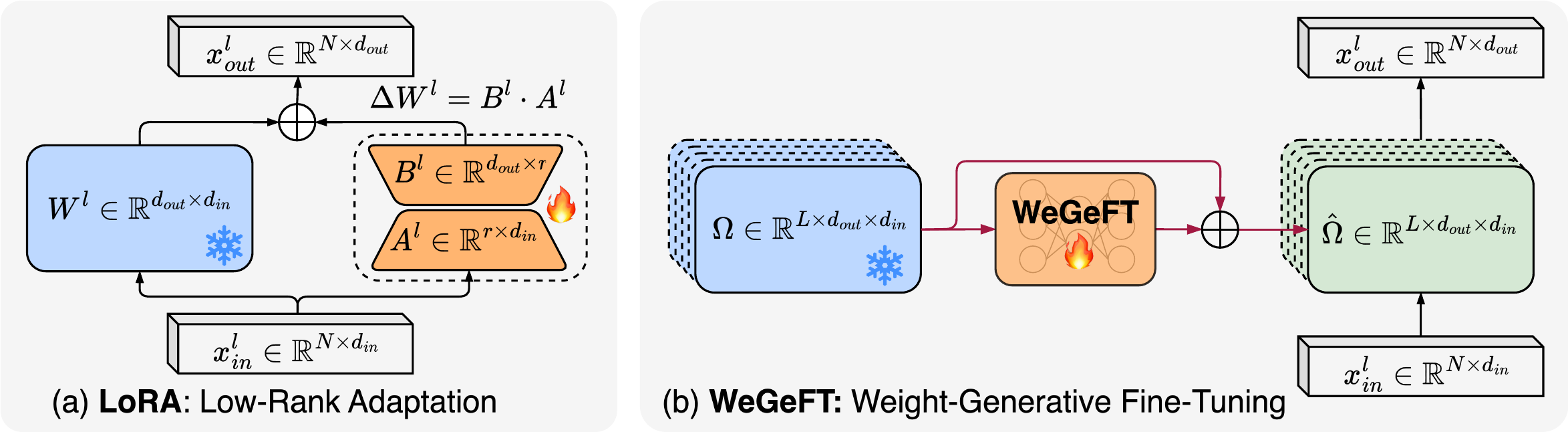}
    \vspace{0.1em}
    \caption{ 
    Comparisons between (a) LoRA~\citep{lora} and (b) our proposed WeGeFT.
    }
    \label{fig:gift-lora}%
\end{figure*}

Low-Rank Adaptation (LoRA) \citep{lora} is a pioneering and widely adopted approach that achieves built-in efficiency in parameters, compute, and memory. LoRA learns fine-tuning weight residuals in low-rank forms for pretrained weights on a layer-specific basis (see the left of Fig. \ref{fig:gift-lora}). Thanks to its strong applicability and promising performance, many follow-up works have emerged, such as DoRA~\citep{dora} and VeRA~\citep{vera}. However, these variants often sacrifice compute efficiency (training wall time) and/or GPU memory efficiency to achieve reductions in learnable parameters or  performance gains on certain downstream tasks. As we demonstrate in experiments, DoRA, while matching or slightly surpassing LoRA’s performance, increases training wall time by more than 5x and consumes around 3GB more GPU memory. On the other hand, VeRA, though significantly reducing the number of learnable parameters, performs much worse than LoRA while drastically increasing training wall time (by more than 20x) and consuming similar additional GPU memory. \textbf{These trade-offs motivate us to seek a formulation that can significantly reduce the number of learnable parameters, achieve superior or on-par performance compared with LoRA, and retain its efficiency in compute and memory.}

Towards these objectives, the recently proposed ReFT \citep{loreft} introduces a promising framework that focuses on lightweight representation-editing modules instead of learning weight residuals, as LoRA does. ReFT is inspired by causal intervention mechanisms \citep{dii} and operates in a layer-specific manner. While ReFT methods reduce learnable parameters and retain compute and memory efficiency comparable to LoRA, they often show inferior performance, as confirmed in our experiments. Additionally, selecting where to intervene within the model to achieve strong downstream task performance is non-trivial. For instance, DiReFT, one of the two ReFT formulations, can be interpreted as applying LoRA directly to hidden representations at specific intervention points. \textbf{This motivates us to seek a unified perspective between parameter-efficient and representation-efficient fine-tuning that enables simpler formulations while achieving on-par or better performance.}

In summary, our experiments reveal the limitations of LoRA variants such as DoRA and VeRA, as well as alternative methods like ReFT, highlighting their drawbacks in compute, memory, and performance trade-offs. \textbf{Outperforming LoRA while maintaining multi-faceted efficiency in parameters, representations, compute, and memory remains a significant challenge.} In this paper, we propose a novel approach to address this challenge.

To clarify the foundation of our proposal, we first review the formulation of LoRA~\cite{lora}. Denote the pretrained weights of a layer  $l \in L$  of a Transformer model by  $W^l \in \mathbb{R}^{d_{out} \times d_{in}}$, and the fine-tuned weights by  $\hat{W}^l \in \mathbb{R}^{d_{out} \times d_{in}}$. LoRA is defined as:
\begin{align}
    \text{LoRA:} \quad & \hat{W}^l = W^l + B^l\cdot A^l,\label{eq:lora}
\end{align}
where $B^l\in \mathbb{R}^{d_{out}\times r}$ and $A^l\in \mathbb{R}^{{r}\times d_{in}}$ are two low-rank matrices representing new learnable model parameters introduced during fine-tuning, and $r$ is the rank ($r\ll \min(d_{in}, d_{out})$). 
Typically,  $B^l$  is initialized to zeros, while  $A^l$  is randomly initialized.

\begin{figure*} [t]
    \centering
   
    \includegraphics[width=1.0\textwidth]{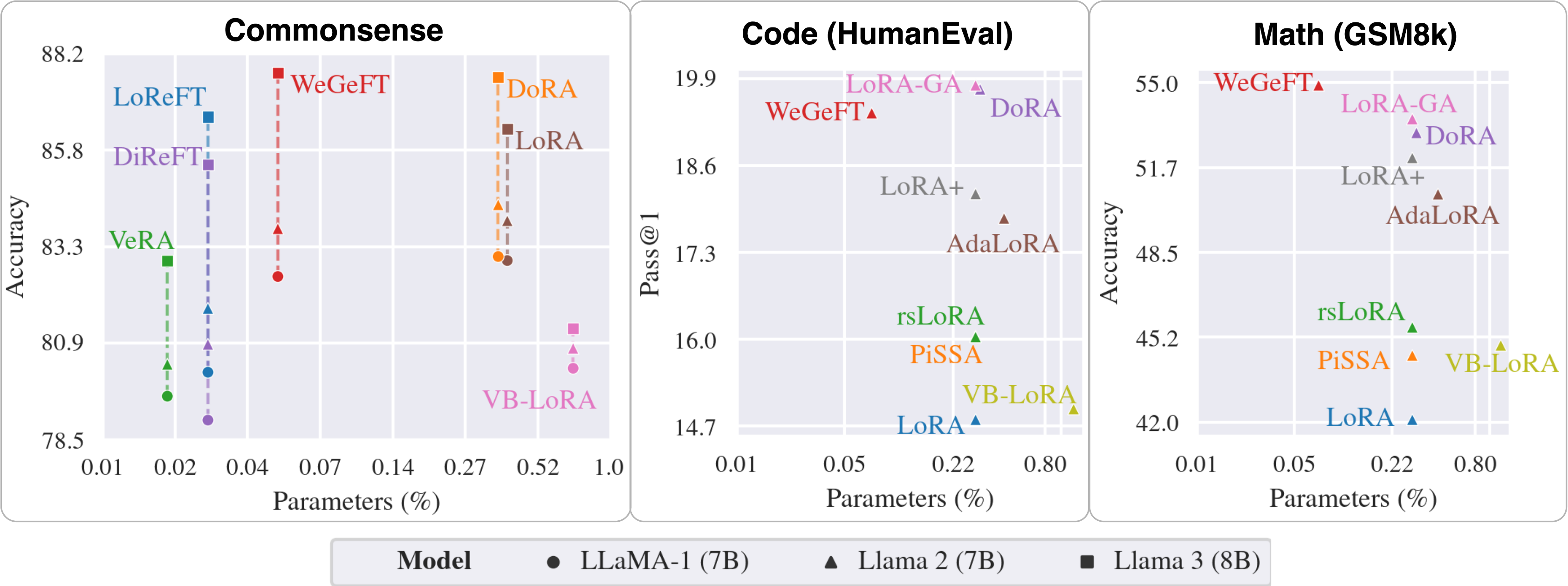}\vspace{1mm}
    \caption{ 
    Comparisons of performance vs. trainable parameters between our WeGeFT  and baseline methods on three tasks using the Llama model family. Figure \ref{fig:multi-faceted-measures} shows that WeGeFT maintains the compute and memory efficiency of LoRA, thus achieving very strong multi-faceted efficiency across parameters, representations, compute and memory. 
    See Section \ref{sec:experiments} for experimental details.
    }
    \label{fig:gift-results}%
\end{figure*}

We rethink LoRA from three key aspects:
\begin{itemize}[leftmargin=*,noitemsep,topsep=0pt]
    \item \textbf{Pretrained Weight Awareness:} LoRA imposes no constraints on  $B^l$  and  $A^l$  beyond their low-rank structure, enabling downstream task data to dictate the fine-tuning process. However, the pretrained weights  $W^l$  encode “carry-over” knowledge that is expected to be useful for downstream tasks. By making fine-tuning weight residuals aware of the pretrained weights, we hypothesize that performance can be further enhanced, especially for stronger pretrained models. Therefore, we aim to parameterize fine-tuning weight residuals in a weight-aware manner.
    \item \textbf{Layer-Specific vs. Shared Adaptation:} In LoRA, $B^l$ and $A^l$ are layer-specific, which ensures layer-local adaptation. Recent methods like Tied LoRA~\citep{tied-lora} propose sharing $ B $ and $ A $ across selected layers, while VB-LoRA~\citep{vblora} introduces a shared vector bank to compose all low-rank matrices via a differentiable top-$ k$ admixture. Although these approaches reduce the number of learnable parameters, our experiments show that they often underperform or significantly degrade performance. For instance, Tied LoRA reduces parameter count but sacrifices expressivity, while VB-LoRA adds unnecessary complexity via top-$k$ modeling. \textit{We advocate for parameter sharing across layers but emphasize that the expressivity of the fine-tuned model should rely on pretrained weights to compensate for the reduction in learnable parameters}. Our results highlight the critical role of pretrained weight awareness in enabling this trade-off.
    \item \textbf{Additive vs. Multiplicative Updates:} LoRA uses additive weight residuals ($B^l \cdot A^l $), which can be limiting in terms of expressivity. Alternatively, multiplicative updates may enable richer, structured transformations. For example, DiReFT~\citep{loreft} applies multiplicative updates in the representation space. Let $y_i^l \in \mathbb{R}^{d_{out} \times 1} $ denote the activation output (representation) for the $ i $-th token at layer $ l $. DiReFT updates it as:
    \begin{align}
    \text{DiReFT:} \quad \hat{y}_i^l &= y_i^l + B^l \cdot (A^l \cdot y_i^l +  b^l), \label{eq:direft} \\ 
    \nonumber &= (\mathbb{I} + B^l \cdot A^l) \cdot y_i^l + B^l \cdot b^l,
    \end{align}
    where $ B^l \in \mathbb{R}^{d_{out} \times r} $, $ A^l \in \mathbb{R}^{r \times d_{out}} $, and $ b^l \in \mathbb{R}^{r \times 1} $ are parameters of the representation-editing module. $\mathbb{I}$ is the identity matrix. While DiReFT introduces multiplicative residuals into the representation space, it is coupled with token intervention search (token-selective).
\end{itemize}

To achieve multi-faceted efficiency across parameters, representations, compute, and memory, we propose \textbf{Weight-Generative Fine-Tuning} (WeGeFT, pronounced \textit{wee-gift}), a simple yet effective formulation (see the right of Fig.~\ref{fig:gift-lora}):
\begin{align}
    \textbf{Our WeGeFT: } \quad  \hat{W}^l &= W^l +  W^l\cdot \phi\cdot \psi,\label{eq:waft_in} \\
    \nonumber &= W^l \cdot (\mathbb{I} + \phi\cdot \psi), 
\end{align}
where $\phi\in \mathbb{R}^{d_{in}\times r}$ and $\psi\in \mathbb{R}^{r\times d_{in}}$ are low-rank matrices shared across layers, and  $r$  is the rank. See Appendix~\ref{sec:gradient} for gradient analyses between LoRA and our WeGeFT.

\begin{figure}%
    \centering
    \includegraphics[width=\linewidth]{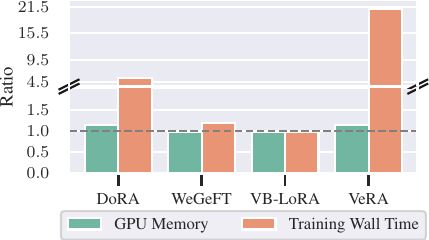}\vspace{1mm}
    \caption{Comparison of the ratio of the GPU Memory (Training Wall Time) for various PEFT methods with the GPU Memory (Training Wall Time) of LoRA. WeGeFT maintains the efficiency of LoRA, as opposed to DoRA and VeRA. Note that while VB-LoRA maintains the memory and compute efficiency, it performs worse than LoRA as seen in Figure \ref{fig:gift-results}.}
    \label{fig:multi-faceted-measures}\vspace{-2mm}
\end{figure}

\begin{itemize}[leftmargin=*,noitemsep,topsep=0pt]
    \item \textbf{Weight-Aware Parameter Sharing:} WeGeFT can be viewed as a weight-aware variant of LoRA, where the layer-specific  $B^l$  in LoRA becomes weight-aware ($B^l = W^l \cdot \phi$) and the layer-specific  $A^l$  becomes layer-agnostic ($A=\psi$). Compared to Tied LoRA~\cite{tied-lora}, WeGeFT retains layer-specific information via  $B^l$, preserving performance. Unlike VB-LoRA~\cite{vblora}, WeGeFT avoids the need for a complex shared vector bank and top-$k$  admixture, instead relying on a pair of shared low-rank matrices  $(\phi, \psi)$  for simplicity and stability. WeGeFT achieves significant improvements in parameter efficiency without sacrificing performance and even enables performance gains when stronger pretrained models (e.g., LLaMA 1 vs. Llama 3) are used.
    \item \textbf{Residual Learning in Weight Space:} WeGeFT extends the residual learning principle of ResNets~\citep{resnet}, $x = x + f(x)$, into the weight space. In contrast, DiReFT~\citep{loreft} applies this principle in the representation space with causal intervention treatments. WeGeFT, however, eliminates the need to search for specific intervention positions and is token-agnostic, ensuring it retains LoRA’s compute and memory efficiency while offering greater flexibility (see Sec.~\ref{sec:bridging-peft-reft}).
\end{itemize}
Fig.~\ref{fig:gift-results} shows result comparisons on three benchmark datasets, which demonstrate the overall multi-faceted efficiency of our proposed WeGeFT. Fig.~\ref{fig:multi-faceted-measures} shows the efficiency comparisons of GPU memory footprint and training wall time. 

\begin{figure*}[t] %
    \centering
    \includegraphics[width=1.0\textwidth]{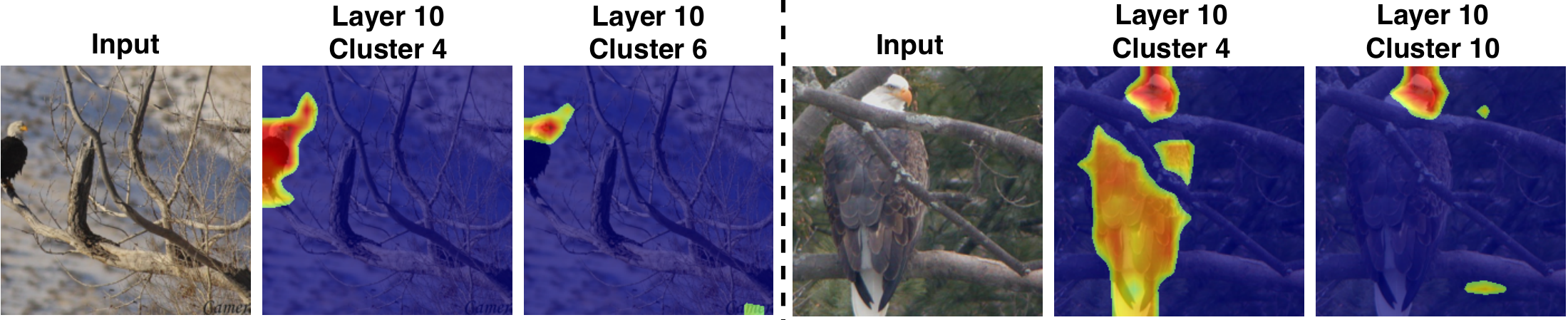} 
    \caption{WeGeFT can play the role of a $r$-way token-clustering head that can localize meaningful objects/parts on images. 
    More examples can be found in Figure \ref{fig:semantic-clusters} in the Appendix.
    }
    \label{fig:cluster-interpretability} %
\end{figure*} 

\begin{itemize}[leftmargin=*,noitemsep,topsep=0pt]
\item \textbf{Visual Inspection of WeGeFT in Computer Vision Tasks}: Let $C^l=W^l\cdot \phi$ be the transformation using the first linear layer of WeGeFT for an output projection layer in MHSA (see Sec.~\ref{sec:visual-classification}). 
We show that $C^l\in \mathbb{R}^{d_{out}\times r}$ can be used as a token-clustering head.
Using the fine-tuned model, the activation of the output projection layer is, $\hat{y}^l \in \mathbb{R}^{ N\times d_{out}}$, where $N$ the number of visual tokens in raster order. 
We compute $r$ heatmaps for visual token clustering by,
\begin{align}
 H^l_{N\times r} = &\hat{y}^l_{N\times d_{out}} \cdot C^l_{d_{out}\times r},
\end{align}
which can highlight semantically meaningful parts of the image.
We normalize the $r$ heatmaps to $[0,1]$ individually and use 0.5 as the threshold in visualizations (Fig.~\ref{fig:cluster-interpretability}).
\end{itemize}

\section{Related Work and Our Contributions}
\label{sec:relatedwork}

\noindent\textbf{Parameter Efficient Fine-tuning (PEFT).} The goal of PEFT methods is to reduce the computational resources (memory footprint, wall time, etc.) required for fine-tuning large models such as Transformers \citep{transformer} and Vision Transformers (ViTs) \citep{vit}. Prompt-based methods either append prompts to the input tokens \citep{prompt-tuning,vpt}, or the intermediate layers \citep{prefix-tuning,prefix-tuning-v2,llama-adapter}. Early work on PEFT used sequential/parallel learnable adapters added after the Multi-Head Self Attention and/or FFN blocks \citep{houlsby-adapter,bapna-adapter,adapter-fusion,invertible-adapters,adapter-drop,compacter,adaptformer}. LoRA \citep{lora} and its variants \citep{adalora,qlora,relora,vera,dft-tune,dora} learn residuals to the pretrained weight matrices in the form of low-rank factorization, removing the added inference cost in adapter based methods. BitFit~\citep{zaken2022bitfit} fine-tunes all the bias terms in a pretrained backbone.
MEND~\citep{mitchellfast} edits a pretrained model by learning fine-tuning weights from the gradient inputs with a low-rank MLP parameterization. FacT \cite{facct} shares the trainable parameters across Self-Attention and MLP blocks. While this reduces trainable parameters, it has a drawback: Transformer modules serve distinct functions - attention layers largely handle syntactical and in-context learning abilities \cite{birth-of-transformer,pruning-heads}, while MLP layers encode factual knowledge \cite{facts-in-gpt}. Sharing parameters across these modules may therefore be suboptimal. Tied-LoRA \citep{tied-lora} shares the residual weights across layers, and also across Query, Key and Value components. In Section \ref{sec:shared-lora-ablations}, we show that the weight-awareness in WeGeFT is essential for parameter sharing across layers, enabling it to outperform Tied-LoRA.

\textbf{Hypernetworks.} \cite{hypernetworks} introduced Hypernetworks, i.e., neural networks that generate the parameters for other neural networks, in language modeling tasks by generating the weights of an LSTM \citep{lstm}. Hypernetworks have previously been applied for few-shot classification \citep{meta-learning-via-hypernetworks,hypertransformer}, transfer learning \citep{requeimaFastFlexibleMultiTask2019} and continual learning \citep{cl-with-hypernetworks,sylph}. Similar to our proposed approach, \citep{requeimaFastFlexibleMultiTask2019} learns to adapt a global feature extractor through an adaptation network. In a few shot continual learning setup, \citep{continual-hypertransformer} uses a hyper-Transformer to generate the parameters for a separate Convolutional Neural Network (ConvNet), which use as inputs both a support set of images of the current task and the ConvNet parameters generated for the previous tasks. HyperFormer++ \citep{hyperformer} uses a Multi-Layer Perceptron (MLP) to generate the parameters from layer embedding and a latent vector for Adapters \citep{houlsby-adapter} introduced across layers of a pretrained model in a multitask setting. Unlike \citep{hyperformer}, we directly use the weights of the frozen pretrained model, thus eliminating the need for embeddings. 

\textbf{Neural Functionals}: Our approach is related to neural functionals that aim to learn deep neural networks acting on the weights of other neural networks. For toy problems, equivariant architectures have been explored for tasks like classifying implicit neural representations \citep{equivariant-deep-weight-spaces,penn, nfn, gnn-equi}, adapting model architectures to new domains \citep{equivariant-deep-weight-spaces}, predicting model generalization performance \citep{penn, nfn, gnn-equi, gnn-meta}, and learned optimizers \citep{unf}. However, our work is the first to explore fine-tuning of a model using it's own weights. We do not use equivariant architectures, but note that this direction of work is orthogonal to ours, and can be further explored in the future.

\textbf{Our Contributions} are summarized as follows: 
\begin{itemize}[leftmargin=*,noitemsep,topsep=0pt]
    \item \textbf{Weight-Generative Fine-Tuning Framework:} WeGeFT introduces a novel formulation where fine-tuning weight residuals are parameterized directly using pretrained weights, leveraging their inherent knowledge for better expressivity, and leading to multiplicative updates in the weight space for richer and more structured transformations compared to additive methods.
    \item \textbf{Multi-Faceted Efficiency:} WeGeFT achieves multi-faceted efficiency across parameters, representation, compute and memory by using shared low-rank matrices $(\phi, \psi)$, significantly reducing learnable parameters while retaining LoRA's simplicity without sacrificing performance (See Figure \ref{fig:gift-results} and Figure \ref{fig:multi-faceted-measures}). 
    \item \textbf{Unified Parameter and Representation Efficiency:} WeGeFT bridges the gap between parameter-efficient and representation-efficient fine-tuning by unifying weight-generative parameterization and shared low-rank matrices.
    \item \textbf{Strong and Scalable Performance:} WeGeFT consistently matches or outperforms LoRA and its variants, achieving superior scalability with stronger pretrained models like Llama 3. 
\end{itemize}

\section{Approach}
In this section, we elaborate on our simple formulation of WeGeFT (Eqn.~\ref{eq:waft_in}) from the more general parameter-generation perspective, which can provide deeper insights.

\subsection{Weight Generation for Explicit Weight-Awareness}
\label{sec:gift-formulation}

For a layer $l\in L$ of a pretrained Transformer model,  to learn its fine-tuning weights, denoted by $\Delta W^l$, that are aware of pretrained weights $W^l$ to ``carry over" their knowledge, a general formulation is to train a generator network, 
\begin{equation}
    \Delta W^l = \mathbb{G}( W^l; \Theta), \forall l\in L ,
\end{equation}
where $\Theta$ is the learnable parameters of the weight generator. 

$\mathbb{G}(\cdot; \Theta)$ needs to be parameterized in a way to meet the desired multi-faceted efficiency. Inspired by the low-rank parameterization scheme in LoRA, we have, 
\begin{equation}
    \mathbb{G}( W^l; \Theta) = f_{\psi} \circ g_{\theta} \circ f_{\phi} (W^l), \forall l\in L , \label{eq:waft-generator}
\end{equation}
where both $f_{\phi}: \mathbb{R}^{d_{out}\times d_{in}}\rightarrow \mathbb{R}^{d_{out}\times r}$ and $f_{\psi}(x):\mathbb{R}^{d_{out}\times r}\rightarrow \mathbb{R}^{d_{out}\times d_{in}}$ are  linear projection layers (without bias terms) with $r$ representing the ``rank". $g_{\theta}: \mathbb{R}^{d_{out}\times r} \rightarrow \mathbb{R}^{d_{out}\times r}$ realizes latent transformations in the low $r$-dim space. $\Theta=(\phi, \psi, \theta)$ collects all learnable parameters.

We note that this design offers a very flexibile way to capture underlying contingency between the fine-tuning weight residuals and the pretrained weights in all $L$ layers. Surprisingly, we observe that we do not need $g_{\theta}$ based on our ablation studies (see Section~\ref{sec:parametrization}). In other words, $g_{\theta}$ is an identity transformation, leading to the simple formulation in Eqn.~\ref{eq:waft_in}. Our (post-hoc) intuitive understanding is:
\begin{itemize}[leftmargin=*,noitemsep,topsep=0pt]
    \item If $g_{\theta}$ includes only linear transformations, it can be naturally absorbed into $f_{\phi}$ and/or $f_{\psi}$. 
    \item If $g_{\theta}$ is an overall non-linear transformation, $\mathbb{G}(W^l; \Theta)$ applies the non-linear transformation in the weight space, which may not be necessary. After all, iterative updates in weight space, including the from-scratch-training of the pretrained weights $W^l$ themselves, are mostly simple updates based on SGD. Nonlinear transformations may be destructive to the ``carry over" knowledge in the pretrained weights, thus negatively impact weight-awareness. More importantly, with a nonlinear $g_{\theta}$, our WeGeFT will sacrifice the multiplicative updates in Eqn.~\ref{eq:waft_in}, $\hat{W}^l = W^l \cdot (\mathbb{I} + \phi\cdot \psi)$, to additive updates, $\hat{W}^l = W^l  + g_{\theta}(W^l \cdot \phi) \cdot \psi$, which will also significantly impact compute and memory efficiency.
\end{itemize}

\textbf{WeGeFT can be applied along the $d_{out}$ dimension too.} In Eqn.~\ref{eq:waft_in}, $(\phi, \psi)$ are applied along the $d_{in}$ dimension of pretrained weights $W^l$.  It is straightforward to apply WeGeFT along the $d_{out}$ dimension by,
    \begin{align}
        \hat{W}^l=W^l + (\phi\cdot \psi)^\top \cdot W_l, \label{eq:gift_dout}         
    \end{align}
where $\phi\in \mathbb{R}^{d_{out}\times r}$ and $\psi\in \mathbb{R}^{r\times d_{out}}$. 

\textbf{WeGeFT Without Parameter Sharing.} It is straightforward to apply our WeGeFT (Eqn.~\ref{eq:waft_in} and Eqn.~\ref{eq:gift_dout}) \textit{without} sharing $(\phi, \psi)$ across layers (denoted by WeGeFT-Sep), which will increase the learnable parameters of the counterpart (WeGeFT with parameter sharing), and to the same as LoRA. We show that WeGeFT-Sep can obtain on-par or better performance than LoRA, demonstrating the advantage of weight-awareness. This flexibility allows WeGeFT to scale more elegantly to larger and diverse datasets, which cannot be achieved by prior methods like VeRA and VB-LoRA.

\subsection{WeGeFT as Token-Agnostic ReFT}
\label{sec:bridging-peft-reft}

 Consider a linear layer with pretrained weights $W^l$ and the pretrained bias term  $b^l$, and WeGeFT weights $\hat{W}^l=(\mathbb{I}+\phi\cdot \psi)\cdot W^l$ (Eqn.~\ref{eq:waft_in}). For an input $x^l$, the output representation/activation at this layer is,
\begin{align}   
   \hat{y}^l &= x^l \cdot \hat{W}^{l^\top} + b^l = \hat{x}^l \cdot W^{l^\top} + b^l,   
    \label{eq:gift_linear_output} \\
   \nonumber \hat{x}^l & = x^l \cdot (\mathbb{I}+\phi\cdot \psi)^\top, 
\end{align}
where $\hat{x}^l$ is the ``fine-tuned" input representation/activation using the same WeGeFT parameters. Hence, \textbf{our WeGeFT can be equivalently applied to the input activation, rather than the pretrained weights, to achieve the same fine-tuning effect}, maintaining the memory and compute efficiency of LoRA in implementation. Unlike the ReFT~\citep{loreft} that entails a dedicated search for where the representation interventions should apply at the token level, our WeGeFT eliminates the need of search, enabling token-agnosticity. Thanks to the parameter sharing, our WeGeFT can retain the representation efficiency.

\section{Experiments}
\label{sec:experiments}

We conduct extensive experiments across Natural Language Generation and Visual Recognition, and compare our two-linear-layer parameterized WeGeFT with various PEFT methods and ReFT. We also conduct ablation studies on the different parameterization schemes of WeGeFT.  More details can be found in Appendix~\ref{sec:impl}. In all the experiments, we follow the baselines in selecting the layers to fine-tune for downstream tasks for fair comparisons. For clarity, we use WeGeFT$_{d_{in}}$ (Eqn.~\ref{eq:waft_in}), WeGeFT$_{d_{out}}$ (Eqn.~\ref{eq:gift_dout}), and WeGeFT-Sep$_{d_{in}}$ (Eqn.~\ref{eq:waft_in} but without parameter sharing).

\begin{table}[t]
\centering
\caption{Results of fine-tuning Llama 2 (7B) on the \textbf{MetaMathQA} dataset and evaluating it on the \textbf{GSM8k {\tt test}} set. 
All baseline results are obtained from \cite{lora-ga}, except VB-LoRA is trained by us together with our proposed WeGeFT. We train our WeGeFT and VB-LoRA by following the same settings in \cite{lora-ga}. }
\label{tab:gsm8k}
\resizebox{0.45\textwidth}{!}{
\begin{tabular}{l|c|c}
\toprule
\textbf{Method} & \textbf{\%Trainable} & \textbf{GSM8k (Acc.)}\\
\midrule
Full & 100 & 54.20\scriptsize{$\pm$0.42} \\
LoRA \cite{lora} & 0.297 & 42.08\scriptsize{$\pm$0.04} \\
PiSSA \cite{pissa} & 0.297 & 44.54\scriptsize{$\pm$0.27} \\
rsLoRA \cite{rs-lora} & 0.297 & 45.62\scriptsize{$\pm$0.10} \\
LoRA+ \cite{lora-plus} & 0.297 & 52.11\scriptsize{$\pm$0.62} \\
DoRA \cite{dora} & 0.317 & 53.07\scriptsize{$\pm$0.75} \\
AdaLoRA \cite{adalora} & 0.445 & 50.72\scriptsize{$\pm$1.39} \\
LoRA-GA \cite{lora-ga} & 0.297 & 53.60\scriptsize{$\pm$0.30} \\
VB-LoRA \cite{vblora} & 1.194 & 44.93\scriptsize{$\pm$1.52} \\
\rowcolor{blue!50}Our WeGeFT$_{d_{in}}$ & \textbf{0.068} & \textbf{54.89}\scriptsize{$\pm$0.92} \\
\bottomrule
\end{tabular}}%
\end{table}

\begin{table}[t]
    \centering
    \caption{Results of fine-tuning  Llama 1 and 2 (7B) on the \textbf{Math10k} benchmark. The Mem. refers to GPU memory, and Wall Time is the time required to complete 1 epoch of training. All results are obtained by us using our code base for fair comparisons, except those by DiReFT and LoReFT  using LLaMA 1 are from~\cite{loreft}.}
    \resizebox{\linewidth}{!}{
    \begin{tabular}{l|l|c|c|c|cccc|c}
        \toprule
        & \textbf{Method} & \rotatebox{90}{\textbf{Params (\%)}} & \rotatebox{90}{\textbf{Mem. (GB)}} & \rotatebox{90}{\textbf{Wall Time}} & \rotatebox{90}{\textbf{AQuA}} & \rotatebox{90}{\textbf{GSM8k}} & \rotatebox{90}{\textbf{MAWPS}} & \rotatebox{90}{\textbf{SVAMP}} & \rotatebox{90}{\textbf{Avg. Acc.}} \\ 
        \midrule
        \multirow{10}{*}{\rotatebox{90}{LLaMA-1 (7B)}} & LoRA$^{r=16}$ & 0.416 & 18.01 & 0.43 & 23.5 & 38.5 & 85.3 & 56.4 & \textbf{50.9} \\
        & DoRA$^{r=16}$ & 0.427 & 20.37 & 2.36 & 21.5 & 37.9 & 86.0 & 55.3 & 50.2 \\
        & \cellcolor{blue!50} WeGeFT-Sep$_{d_{in}}$ & \cellcolor{blue!50} 0.416 & \cellcolor{blue!50}18.01 & \cellcolor{blue!50}0.46 & \cellcolor{blue!50}23.8 & \cellcolor{blue!50}37.9 & \cellcolor{blue!50}84.5 & \cellcolor{blue!50}54.2 & \cellcolor{blue!50}50.1 \\ \cmidrule{2-10}
        & LoRA$^{r=2}$ & 0.052 & 17.74 & 0.43 & 23.1 & 34.6 & 83.9 & 54.1 & 48.9 \\
        & DoRA$^{r=2}$ & 0.065 & 20.09 & 2.36 & 21.1 & 34.6 & 84.0 & 53.8 & 48.4 \\
        & VeRA & 0.042 & 20.65 & 9.01 & 21.3 & 34.0 & 82.8 & 50.7 & 47.2 \\
        & FacT-TT & 0.051 & 17.74 & 0.52 & 21.5 & 30.7 & 80.3 & 50.3 & 45.7 \\
        & FacT-TK & 0.062 & 17.75 & 0.59 & 21.3 & 34.8 & 82.2 & 51.9 & 47.5 \\
        & VB-LoRA & 0.840 & 18.33 & 0.42 & 21.26 & 29.3 & 78.9 & 49.5 & 44.7 \\
         & \cellcolor{blue!50} WeGeFT$_{d_{in}}$ & \cellcolor{blue!50}0.052 & \cellcolor{blue!50}17.74 & \cellcolor{blue!50}0.51 & \cellcolor{blue!50}24.3 & \cellcolor{blue!50}36.5 & \cellcolor{blue!50}82.4 & \cellcolor{blue!50}56.9 & \cellcolor{blue!50}\textbf{50.0} \\ \cmidrule{2-10}
        & DiReFT & 0.031 & 31.42 & 0.26 & 21.3 & 24.1 & 74.5 & 42.7 & 40.6 \\
        & LoReFT & 0.031 & 55.42 & 0.29 & 21.4 & 26.0 & 76.2 & 46.8 & 42.6 \\
        & \cellcolor{blue!50} WeGeFT$_{d_{out}}$ & \cellcolor{blue!50}0.016 & \cellcolor{blue!50}17.71 & \cellcolor{blue!50}0.36 & \cellcolor{blue!50}20.74 & \cellcolor{blue!50}33.0 & \cellcolor{blue!50}80.8 & \cellcolor{blue!50}53.5 & \cellcolor{blue!50}\textbf{47.0} \\
        \midrule
        \multirow{10}{*}{\rotatebox{90}{Llama 2 (7B)}} & LoRA$^{r=16}$ & 0.416 & 18.01 & 0.43 & 24.5 & 43.4 & 86.1 & 57.2 & 52.8 \\
        & DoRA$^{r=16}$ & 0.429 & 20.37 & 2.35 & 24.1 & 41.4 & 87.1 & 57.1 & 52.4 \\
        & \cellcolor{blue!50} WeGeFT-Sep$_{d_{in}}$ & \cellcolor{blue!50}0.416 & \cellcolor{blue!50}18.01 & \cellcolor{blue!50}0.46 & \cellcolor{blue!50}26.1 & \cellcolor{blue!50}42.4 & \cellcolor{blue!50}85.9 & \cellcolor{blue!50}58.6 & \cellcolor{blue!50}\textbf{53.3} \\ \cmidrule{2-10}
        & LoRA$^{r=2}$ & 0.052 & 17.74 & 0.42 & 24.7 & 40.2 & 85.0 & 56.0 & 51.5 \\
        & DoRA$^{r=2}$ & 0.065 & 20.09 & 2.35 & 24.0 & 40.6 & 84.6 & 56.0 & 51.3 \\
        & VeRA & 0.042 & 20.65 & 9.00 & 23.5 & 38.7 & 85.3 & 54.3 & 50.4 \\
        & VB-LoRA & 0.840 & 18.33 & 0.43 & 22.4 & 33.4 & 81.4 & 52.4 & 47.4 \\
        & FacT-TT & 0.051 & 17.74 & 0.52 & 24.9 & 38.3 & 81.9 & 56.2 & 50.3 \\
        & FacT-TK & 0.062 & 17.75 & 0.59 & 24.5 & 41.0 & 85.7 & 54.4 & 51.4 \\
         & \cellcolor{blue!50} WeGeFT$_{d_{in}}$ & \cellcolor{blue!50}0.052 & \cellcolor{blue!50}17.74 & \cellcolor{blue!50}0.50 & \cellcolor{blue!50}23.6 & \cellcolor{blue!50}42.4 & \cellcolor{blue!50}84.2 & \cellcolor{blue!50}57.4 & \cellcolor{blue!50}\textbf{51.9} \\ \cmidrule{2-10}
        & DiReFT & 0.031 & 31.42 & 0.26 & 20.5 & 27.9 & 77.5 & 45.9 & 42.9 \\
        & LoReFT  & 0.031 & 55.42 & 0.29 & 24.8 & 31.7 & 79.6 & 50.9 & 46.7 \\
        & \cellcolor{blue!50} WeGeFT$_{d_{out}}$ & \cellcolor{blue!50}0.016 & \cellcolor{blue!50}17.71 & \cellcolor{blue!50}0.39 & \cellcolor{blue!50}26.1 & \cellcolor{blue!50}38.0 & \cellcolor{blue!50}83.1 & \cellcolor{blue!50}57.3 & \cellcolor{blue!50}\textbf{51.1} \\
        \bottomrule
    \end{tabular}
    }
    \label{tab:math-reasoning} \vspace{-4mm}
\end{table}

\subsection{Arithmetic Reasoning}\label{sec:arithmetic}
We demonstrate the multi-faceted efficiency of WeGeFT  with experiments on the Math10k benchmark \cite{llmadapters} for Arithmetic Reasoning, which is a small scale dataset enabling comprehensive evaluations. We conduct further experiments with MetaMathQA \cite{metamathqa}, a larger and higher quality fine-tuning dataset, to understand the impact of higher quality data on WeGeFT.  In both experiments, we evaluate the model on the final answer following the same protocol used in prior works. Experimental details and hyperparameters are provided in Appendix \ref{sec:math-hyperparameters}.

\textbf{Results}: Table \ref{tab:gsm8k} shows the results of finetuning using the  MetaMathQA datasets, and evaluating on the GSM8k test set. WeGeFT outperforms all the prior methods with 4 times fewer parameters, and slightly outperform the full fine-tuning. We use the relatively smaller Math10k benchmark (Table \ref{tab:math-reasoning}) to run comprehensive evaluations for WeGeFT with respect to multi-faceted efficiency.

\textbf{\textbullet\ Parameter Efficiency}: WeGeFT efficiently adapts both strong and weak models, as shown by finetuning LLaMA-1 and LLaMA-2 (7B) on the Math10k benchmark in Table \ref{tab:math-reasoning}. LLaMA-1 (7B), a weak model for arithmetic reasoning (11\% zero-shot accuracy on GSM8k \cite{llama}), requires substantial adaptation. Table \ref{tab:math-reasoning} shows WeGeFT adapts LLaMA-1 (7B) far more effectively than prior PEFT methods. Reducing parameters in LoRA and DoRA significantly degrades performance—LoRA's accuracy drops from 50.9 ($r=16$) to 48.9 ($r=2$). At a comparable parameter count (~0.05\%), WeGeFT outperforms all baselines (LoRA, DoRA, VeRA, VB-LoRA) and nearly matches LoRA at $r=16$. \textbf{This demonstrates that generating fine-tuning residuals from pretrained weights improves parameter efficiency by enabling adaptation with minimal parameters, whereas LoRA and DoRA with ${r=2}$ struggle to perform well}.

\textbf{\textbullet\ Computational and Memory Efficiency}:  \textit{WeGeFT maintains compute and memory efficiency while achieving equal or higher accuracy than prior PEFT methods, unlike DoRA and VeRA, which compromise efficiency}. Although VeRA reduces trainable parameters, it requires a large intermediate dimension for fixed random weights (12288 here). On Math10k, VeRA takes $\sim$9 hours/epoch and 20.65GB GPU memory, whereas LoRA and WeGeFT need only $\sim$0.5 hours/epoch and 17.74GB under the same setup. Table \ref{tab:training-time-analysis} in Appendix \ref{sec:training-time-analysis} confirms that reducing the intermediate dimension to 1024 (as in \cite{vera}) lowers time and memory costs but causes a severe accuracy drop. Table \ref{tab:float16-analysis} in Appendix \ref{sec:float16-analysis} shows that WeGeFT is efficient even with mixed-precision float16. Thus, Similar to LoRA, WeGeFT can be used even with consumer GPUs.

\begin{table*}[t]
    
    \centering
    \caption{Results of fine-tuning LLaMA-1 7B, Llama 2 7B and Llama 3 8B on eight Commonsense Reasoning benchmarks (\textbf{Commonsense170k}). ReFT results are obtained from \cite{loreft}.}
    \resizebox{0.95\textwidth}{!}{
    \begin{tabular}{ll|c| c c c c c c c c|c}
        \toprule
        & \textbf{Method} & \textbf{Params (\%)} & \textbf{BoolQ} & \textbf{PIQA} & \textbf{SIQA} & \textbf{HellaS.} & \textbf{WinoG.} & \textbf{ARC-e} & \textbf{ARC-c} & \textbf{OBQA} & \textbf{Avg} \\ \midrule
        \multirow{8}{*}{\rotatebox{90}{LLaMA-1 (7B)}} & 
        LoRA~\citep{lora} & 0.416 & 73.3 & 84.5 & 80.4 & 94.2 & 85.5 & 87.6 & 72.6 & 85.6 & 83.0 \\
        & DoRA~\citep{dora} & 0.427 & 73.4 & 84.8 & 80.7 & 94.4 & 85.7 & 87.4 & 72.4 & 85.9 & \textbf{83.1} \\
        & {VeRA}~\citep{vera} & {0.023} & {70.4} & {82.4} & {79.9} & {91.4} & {81.8} & {83.3} & {67.0} & {80.6} & {79.6} \\
        & VB-LoRA~\citep{vblora} & 0.840 & 70.5 & 82.6 & 79.3 & 92.5 & 83.1 & 84.5 & 68.1 & 81.7 & 80.3 \\
          & \cellcolor{blue!50}WeGeFT$_{d_{in}}$ & \cellcolor{blue!50}0.052 & \cellcolor{blue!50}72.8 & \cellcolor{blue!50}84.7 & \cellcolor{blue!50}80.8 & \cellcolor{blue!50}93.9 & \cellcolor{blue!50}84.3 & \cellcolor{blue!50}86.7 & \cellcolor{blue!50}72.3 & \cellcolor{blue!50}85.1 & \cellcolor{blue!50}82.6 \\ \cmidrule{2-12}
        & DiReFT~\citep{loreft} & 0.031 & 69.5 & 83.0 & 79.0 & 92.5 & 80.5 & 82.2 & 68.0 & 77.5 & 79.0 \\
        & LoReFT~\citep{loreft} & 0.031 & 69.3 & 84.4 & 80.3 & 93.1 & 84.2 & 83.2 & 68.2 & 78.9 & 80.2 \\
          & \cellcolor{blue!50} WeGeFT$_{d_{out}}$  & \cellcolor{blue!50}0.016 & \cellcolor{blue!50}71.5 & \cellcolor{blue!50}83.4 & \cellcolor{blue!50}81.1 & \cellcolor{blue!50}93.6 & \cellcolor{blue!50}83.7 & \cellcolor{blue!50}86.1 & \cellcolor{blue!50}72.0 & \cellcolor{blue!50}83.9 & \cellcolor{blue!50}\textbf{81.9} \\
        \midrule
        \multirow{9}{*}{\rotatebox{90}{Llama 2 (7B)}}  & LoRA~\citep{lora}  & 0.416 & 74.8 & 85.9 & 80.8 & 94.8 & 86.3 & 88.3 & 75.4 & 85.9 & 84.0 \\ 
        & DoRA~\citep{dora}  & 0.427 & 74.6 & 86.2 & 81.1 & 94.9 & 86.8 & 89.1 & 75.9 & 86.7 & \textbf{84.4} \\
        & {VeRA}~\citep{vera} & {0.023} & {71.9} & {82.2} & {80.0} & {92.2} & {83.3} & {84.3} & {68.8} & {80.5} & {80.4} \\
        & VB-LoRA~\citep{vblora} & 0.840 & 71.6 & 83.2 & 79.7 & 92.8 & 83.5 & 84.8 & 69.3 & 81.9 & 80.8 \\
        & \cellcolor{blue!50} WeGeFT$_{d_{in}}$   & \cellcolor{blue!50}0.052 & \cellcolor{blue!50}73.9 & \cellcolor{blue!50}85.7 & \cellcolor{blue!50}82.1 & \cellcolor{blue!50}94.6 & \cellcolor{blue!50}85.6 & \cellcolor{blue!50}88.3 & \cellcolor{blue!50}74.7 & \cellcolor{blue!50}85.4 & \cellcolor{blue!50}83.8 \\ \cmidrule{2-12}
        & DiReFT~\citep{loreft} & 0.031 & 70.8 & 83.6 & 80.2 & 93.6 & 82.1 & 84.8 & 70.4 & 81.5 & 80.9 \\
    & LoReFT~\citep{loreft} & 0.031 & 71.1 & 83.8 & 80.8 & 94.3 & 84.5 & 85.6 & 72.2 & 82.3 & 81.8 \\ 
          & \cellcolor{blue!50} WeGeFT$_{d_{out}}$ & \cellcolor{blue!50}0.016 & \cellcolor{blue!50}73.4 & \cellcolor{blue!50}85.2 & \cellcolor{blue!50}81.8 & \cellcolor{blue!50}94.3 & \cellcolor{blue!50}85.3 & \cellcolor{blue!50}87.7 & \cellcolor{blue!50}74.9 & \cellcolor{blue!50}83.8 & \cellcolor{blue!50}\textbf{83.3} \\
        \midrule
        \multirow{9}{*}{\rotatebox{90}{Llama 3 (8B)}}  & LoRA~\citep{lora}  & 0.392 & 74.6 & 89.4 & 81.3 & 95.9 & 87.7 & 91.9 & 81.7 & 87.7 & 86.3 \\ 
        & DoRA~\citep{dora}  & 0.361 & 76.2 & 90.8 & 82.1 & 96.7 & 89.0 & 93.5 & 83.4 & 89.1 & 87.6 \\
        & {VeRA}~\citep{vera} & {0.018} & {71.6} & {85.7} & {80.7} & {93.8} & {85.2} & {87.6} & {75.6} & {84.1} & {83.0} \\
        & VB-LoRA~\citep{vblora} & 0.712 & 72.3 & 83.7 & 80.0 & 93.1 & 84.1 & 85.1 & 70.5 & 81.9 & 81.3 \\
         & \cellcolor{blue!50}WeGeFT$_{d_{in}}$  & \cellcolor{blue!50}0.049 & \cellcolor{blue!50}76.0 & \cellcolor{blue!50}89.7 & \cellcolor{blue!50}83.1 & \cellcolor{blue!50}96.7 & \cellcolor{blue!50}89.1 & \cellcolor{blue!50}93.0 & \cellcolor{blue!50}84.4 & \cellcolor{blue!50}89.8 & \cellcolor{blue!50}\textbf{87.7} \\ \cmidrule{2-12}
        & DiReFT~\citep{loreft} & 0.026 & 73.4 & 88.7 & 81.0 & 95.6 & 85.5 & 91.8 & 81.8 & 85.4 & 85.4 \\
    & LoReFT~\citep{loreft} & 0.026 & 75.1 & 90.2 & 82.0 & 96.3 & 87.4 & 92.4 & 81.6 & 87.5 & 86.6 \\         
          & \cellcolor{blue!50}WeGeFT$_{d_{out}}$   & \cellcolor{blue!50}0.013 & \cellcolor{blue!50}75.7 & \cellcolor{blue!50}89.9 & \cellcolor{blue!50}82.5 & \cellcolor{blue!50}96.4 & \cellcolor{blue!50}88.7 & \cellcolor{blue!50}92.5 & \cellcolor{blue!50}82.3 & \cellcolor{blue!50}86.3 & \cellcolor{blue!50}\textbf{86.8} \\
        \bottomrule
    \end{tabular}}
    \label{tab:commonsense-reasoning}\vspace{-2mm}
\end{table*}

\textbf{\textbullet\ Impact of Data Quality}: WeGeFT ``reacts" positively to the quality of training datasets, as can be seen by the relative performance improvement from Math10k benchmark (Table \ref{tab:math-reasoning}) to the MetaMathQA benchmark (Table \ref{tab:gsm8k}).

\textbf{\textbullet\ Comparison with ReFT}: To compare with DiReFT and LoReFT, we apply WeGeFT$_{d_{out}}$ following their fine-tuning strategy (Eqn.~\ref{eq:direft}). WeGeFT achieves higher average accuracy than both ReFT variants while using half the parameters on LLaMA-1 and LLaMA-2. ReFT's higher memory usage (and lower wall time) is due to its implementation lacking gradient checkpointing, whereas our HuggingFace PEFT-based implementation uses checkpointing for better scalability with large models.

\subsection{Commonsense Reasoning} \label{sec:commonsense}
We use combined training data of eight benchmarks (i.e., \textbf{Commonsense170k}, containing a total of 170k training samples), and evaluate on their test sets individually, following the same protocol used in \cite{llmadapters} and \cite{loreft}. 
The examples in the Commonsense170k are formulated as multiple choice questions and consist of BoolQ \citep{boolq}, PIQA \citep{piqa}, SIQA \citep{siqa}, HellaSwag \citep{hellaswag}, WinoGrande \citep{winogrande}, Arc-e and Arc-c \citep{arc}, and OBQA \citep{obqa} datasets. 
We experiment with LLaMA-1 (7B), Llama 2 (7B) and Llama 3 (8B) models. Experimental details including hyperparameters are provided in Appendix \ref{sec:commonsense-hyperparameters}.

\textbf{Results} are shown in Table \ref{tab:commonsense-reasoning}. Based on the observations of our initial experiments, WeGeFT-Sep  does not show significant improvement over WeGeFT. Hence, we focus on evaluating WeGeFT on the Commonsense170k benchmark. We summarize the observations as follows:
\begin{itemize}[leftmargin=*,noitemsep,topsep=0pt]
    \item \textbf{The weight-awareness of our WeGeFT is positively correlated with the expressivity of the pretrained models.} WeGeFT slightly outperforms DoRA in fine-tuning Llama 3 (8B) using 8x fewer parameters, while DoRA outperforms WeGeFT in fine-tuning both LLaMA 1 and Llama 2 at the expense of 8x more parameters and more expensive training cost. Considering the  trend of increasingly powerful pretrained large foundation models, WeGeFT shows a very promising potential due to its efficiency and strong performance.  
    \item Although VeRA uses less parameters, its performance is much worse and the training cost is very high, similar to the observations on Math10k. Furthermore, our WeGeFT$_{d_{out}}$ outperforms VeRA with even less parameters and much more efficient training. 
    \item Compared with DiReFT and LoReFT, our WeGeFT is still better and reduces the parameters by half. 
\end{itemize}

\subsection{Instruction Following} \label{sec:instruction_tuning}

\begin{table}[t]
\centering
\caption{Results of fine-tuning Llama 2 (7B) on the \textbf{WizardLM} dataset \cite{wizardlm} and evaluating it on the \textbf{MT-Bench} \cite{mt-bench}. All the results except VB-LoRA are obtained from \cite{lora-ga}. We train VB-LoRA and WeGeFT following the same settings as \cite{lora-ga}.}
\label{tab:mtbench}
\resizebox{0.99\linewidth}{!}{
\begin{tabular}{l|c|c}
\toprule
\textbf{Method} & \textbf{Params (\%)} &  \textbf{First Turn Score}\\
\midrule
Full & 100 & 5.56\scriptsize{$\pm$0.09}\\
LoRA \cite{lora} & 0.297 & 5.61\scriptsize{$\pm$0.10}\\
PiSSA \cite{pissa} & 0.297 & 5.30\scriptsize{$\pm$0.02}\\
rsLoRA \cite{rs-lora} & 0.297 & 5.25\scriptsize{$\pm$0.03}\\
LoRA+ \cite{lora-plus} & 0.297 & 5.71\scriptsize{$\pm$0.08}\\
DoRA \cite{dora} & 0.317 & \textbf{5.97}\scriptsize{$\pm$0.02}\\
AdaLoRA \cite{adalora} & 0.445 & 5.57\scriptsize{$\pm$0.05}\\
LoRA-GA \cite{lora-ga} & 0.297 & 5.95\scriptsize{$\pm$0.16}\\
VB-LoRA \cite{vblora} & 1.194 & 5.57\scriptsize{$\pm$0.05}\\
\rowcolor{blue!50}WeGeFT$_{d_{in}}$ & \textbf{0.068} & 5.75\scriptsize{$\pm$0.11}\\
\bottomrule
\end{tabular}}\vspace{-2mm}
\end{table}

We fine-tune LLaMA-2 (7B) on a 52k subset of the WizardLM dataset \cite{wizardlm}, filtering out samples containing ``Sorry'' and ``As an AI'' following \cite{lora-ga}. We evaluate the instruction following ability on the MT-Bench dataset \cite{mt-bench}, which spans domains such as math, roleplay, reasoning, and coding. We report the the single turn score by prompting an LLM judge (\textbf{GPT4}) to rate the responses from the fine-tuned model from 1-10. Experimental settings and hyperparameters can be found in Appendix \ref{sec:instruct-hyperparameters}).

\textbf{Results} are shown in Table \ref{tab:mtbench}. Our WeGeFT is on-par with LoRA-GA and DoRA with 4 times fewer parameters. It also slightly outperforms the full fine-tuning.

\subsection{Code Generation}
\label{sec:code-generation}
We fine-tune Llama 2 (7B) using the Code-Feedback dataset \cite{code-feedback}, which is a multi-turn dataset containing execution and human feedback. We evaluate the fine-tuned models on HumanEval \cite{human-eval}, containing Python problems evaluated for functional correctness. Experimental settings and hyperparameters are in Section \ref{sec:code-hyperparameters}.

\begin{table}[h]
\centering
\caption{Results of fine-tuning Llama 2 (7B) on the \textbf{Code-Feedback} dataset \cite{code-feedback} and evaluating  it on the \textbf{HumanEval} \cite{human-eval}. All results except VB-LoRA are obtained from \cite{lora-ga}. We train VB-LoRA and WeGeFT following the same settings as \cite{lora-ga}.}
\label{tab:humaneval}
\resizebox{0.99\linewidth}{!}{
\begin{tabular}{l|c|c}
\toprule
\textbf{Method} & \textbf{Params} (\%) & \textbf{Pass@1}\\
\midrule
Full & 100 &\textbf{19.87}\scriptsize{$\pm$0.57}\\
LoRA \cite{lora} & 0.297 & 14.76\scriptsize{$\pm$0.17}\\
PiSSA \cite{pissa} & 0.297 & 16.02\scriptsize{$\pm$0.78}\\
rsLoRA \cite{rs-lora} & 0.297 & 16.01\scriptsize{$\pm$0.79}\\
LoRA+ \cite{lora-plus} & 0.297 & 18.17\scriptsize{$\pm$0.52}\\
DoRA \cite{dora} & 0.317 & 19.75\scriptsize{$\pm$0.41}\\
AdaLoRA \cite{adalora} & 0.445 & 17.80\scriptsize{$\pm$0.44}\\
LoRA-GA \cite{lora-ga} & 0.297 & 19.81\scriptsize{$\pm$1.46}\\
VB-LoRA \cite{vblora} & 1.194 & 14.92\scriptsize{$\pm$0.92}\\
\rowcolor{blue!50}WeGeFT$_{d_{in}}$ & \textbf{0.068} & 19.39\scriptsize{$\pm$0.68}\\
\bottomrule
\end{tabular}}%
\end{table}

\textbf{Results} are shown in Table \ref{tab:humaneval}. Our WeGeFT is on-par with LoRA-GA and DoRA.

\subsection{Visual Recognition}
\label{sec:visual-classification}

\textbf{Data.} We evaluate WeGeFT on the VTAB-1k benchmark \cite{vtab} and the fine-grained visual classification (FGVC) benchmark containing Caltech-UCSD Birds \citep{cub}, NABirds \citep{nabirds}, Oxford Flowers \citep{flowers}, Stanford Cars \citep{cars}, and Stanford Dogs \citep{dogs}. 

\textbf{Models.} We use the ViT-B/16 architecture \citep{vit} pretrained on ImageNet21k dataset \citep{imagenet21k} using a supervised objective, with the checkpoints from the {\tt timm} package~\citep{rw2019timm}.  
We apply LoRA and WeGeFT to the output projection layers in MHSA, which is inspired by observations in~\citep{artihippo}. All hyperparameters are provided in Appendix~\ref{sec:impl-vision}.

\textbf{Results:} Tables \ref{tab:fgvc-results} and \ref{tab:vtab-results} show that our WeGeFT performs better than other PEFT methods on both FGVC, while using fewer parameters. The GPU memory consumption is similar among the different methods with negligible differences. With 5.9 times less parameters used (0.025M {\tt vs} 0.147M), on FGVC tasks, our WeGeFT improves LoRA by 0.68\% Top-1 accuracy. 

\begin{table}[!h]
    \centering
    \caption{Results on the finegrained visual classification (FGVC) tasks with ViT-B/16 pretrained on ImageNet21k. The number of trainable parameters are reported without the classification head (which has the same number of parameters for all the methods).
    }
    \resizebox{\linewidth}{!}{
    \begin{tabular}{l|l|lrrrr|r}
\toprule
      \textbf{Method} &  \textbf{Params (M)} &  \textbf{CUBS} &  \textbf{Bird} &  \textbf{Flower} &   \textbf{Dog} &   \textbf{Car} &  \textbf{Avg} \\
        \midrule
    VPT      &         0.046  &                    87.88 &   84.79 &     98.98 &  84.51 &  82.89 &  87.81 \\
     BitFit     &         0.083  &                    87.75 &   84.61 &     \textbf{99.32} &  85.23 &  84.01 &  88.18 \\
LoRA &         0.147  &                    88.00 &   84.94 &     \textbf{99.32} &  85.36 &  \textbf{85.92} &  88.71 \\
\rowcolor{blue!50} WeGeFT$_{d_{in}}$ &         \textbf{0.025}  &                    \textbf{89.71} &   \textbf{86.28} &     99.22 &  \textbf{87.44} &  84.28 &  \textbf{89.39} \\
\bottomrule
\end{tabular}

}%
\label{tab:fgvc-results} \vspace{-3mm}
\end{table}

\begin{table}[!h]
    \centering
    \caption{Results on the VTAB benchmark \cite{vtab} with ViT-B/16 pretrained on ImageNet21k. Trainable parameters are reported the same way as Table \ref{tab:fgvc-results}.}
    \resizebox{\linewidth}{!}{
    \begin{tabular}{c|c|c|c|c|c}
        \toprule
         \textbf{Method} & \textbf{Params (M)} & \textbf{Natural} & \textbf{Specialized} & \textbf{Structured} & \textbf{Avg}  \\
         \toprule
         VPT & 0.046 & 81.0 & 85.7 & 58.9 & 72.7 \\
         BitFit & 0.083 & 81.8 & 85.2 & 57.8 & 72.4 \\
         LoRA & 0.147 & \textbf{82.0} & 85.9 & 61.0 & 74.0 \\
         FacT-TT & 0.040 & 79.8 & 86.0 & 58.0 & 71.9 \\
         FacT-TK & 0.069 & 80.0 & \textbf{86.8} & 60.9 & 73.4 \\
         \rowcolor{blue!50} WeGeFT$_{d_{in}}$ & 0.025 & \textbf{82.0} & 86.3 & \textbf{61.1} & \textbf{74.1} \\
         \bottomrule
    \end{tabular}}
    
    \label{tab:vtab-results}
\end{table}

\section{Ablation Studies}\label{sec:ablation}

\subsection{Different Parameterization Schema for WeGeFT}
\label{sec:parametrization}
As mentioned in Section \ref{sec:gift-formulation}, a simple linear transformation of the pretrained weights works surprisingly well in generating fine-tuning residual weights. To verify effects of non-linear $g_{\theta}()$ in Eqn.~\ref{eq:waft-generator}. We compare,  
\begin{itemize}[leftmargin=*,noitemsep,topsep=0pt]
    \item \textit{Transformer}: We treat the set of shared pretrained weights across $L$ layers as a batch of $L$ sequences each consisting of $d_{out}$ tokens in $r$-dim space (after the first linear project layer $f_{\phi}$), denoted by $\mathcal{W}$. We then apply a single Transformer block~\citep{transformer}. 
    \item \textit{MLP-Mixers}: Similar to vanilla Transformers, we apply a single MLP-Mixer~\citep{tolstikhin2021mlp} block.
    \item \textit{Multi-Layer Perceptrons (MLPs)}: e.g., $g(\mathcal{W};\theta)=\text{Linear}(\text{GELU}(\text{Linear}(\mathcal{W};\theta_1)); \theta_2)$, where $\theta_1\in \mathbb{R}^{m\cdot r\times r +m\cdot r}$ and $\theta_2\in \mathbb{R}^{r\times m\cdot r + r}$ consist of weights and bias terms of the two linear layers with the MLP latent dimension ratio $m$ (e.g., $m=2$).
    \item \textit{Element-wise non-linearity functions} without learnable parameters (i.e., $\theta=\emptyset$): e.g.,  $g(\mathcal{W})=\text{Sigmoid}(\mathcal{W})$ or $g(\mathcal{W})=\text{GELU}(\mathcal{W})$.
\end{itemize}

\begin{table}[!h]
    \centering
    \caption{Comparisons between various non-linear transformations for $g_\theta$ on the FGVC benchmark. 
}
    \resizebox{\linewidth}{!}{
    \begin{tabular}{l|l|lrrrr|r}
\toprule
Schema &  Params (M) &  CUBS &  Bird &  Flower &   Dog &   Car &  \textit{Avg} \\
\midrule
\rowcolor{blue!50} Identity & 0.025 & \textbf{89.71} & \textbf{86.28} & 99.22 & \textbf{87.44} & 84.28 &  \textbf{89.39} \\
Sigmoid & 0.025 & 89.56 & 84.61 & 99.20 & 86.69 & 84.04 &  88.82 \\
GeLU & 0.025 & 89.70 & 85.30 & 99.19 & 86.71 & 83.81 & 88.94 \\
MLP & 0.036 & 89.06 & 85.44 & \textbf{99.30} & 86.17 & 84.24 & 88.84 \\
Transformer & 0.027 & 89.56 & 86.23 & 99.24 & 86.31 & 84.26 & 89.12 \\
MLP Mixer & 0.125 & 88.76 & 86.21 & 99.25 & 86.35 & \textbf{85.66} & 89.25 \\
\bottomrule
\end{tabular}
}
\label{tab:fgvc-schema}\vspace{-4mm}
\end{table}

Through ablation studies on the FGVC benchmark, we verify that using any non-linear transformation for $g_\theta$ results in degradation in performance. We use the same settings as Section \ref{sec:visual-classification}. As seen from Table \ref{tab:fgvc-schema}, the simple two-linear layer formulation achieves better or equivalent performance than all other schema at a lower parameter cost. While we do not have a theoretical understanding yet, we hypothesize that the superior performance of the identity operation over more complex and non-linear operations is because of difficulty in optimization. We only study the non-linear formulation on small models on simpler tasks due to computational constraints, and note that this presents an interesting avenue for future research.

\vspace{-2mm}
\subsection{Alternative Formulation of Tied LoRA}
\label{sec:shared-lora-ablations}
Tied LoRA~\cite{tied-lora} uses a sophisticated design of sharing weights across layers. We test a straightforward parameter sharing  LoRA, i.e., $\Delta W^l = B \cdot A\ \ \forall l \in L$, where $(B,A)$ is shared across layers. 
Table \ref{tab:commonsense-reasoning-shared-lora} shows that this strategy leads to much lower performance than our WeGeFT, which justifies the advantage of weight-awareness.  

\begin{table}[h]
    \centering
    \caption{Comparisons of Shared LoRA and WeGeFT on eight commonsense reasoning benchmarks.}
    \resizebox{0.8\linewidth}{!}{
    \begin{tabular}{ll|c|c}
        \toprule
        & \textbf{Method} & \textbf{Params (\%)} &  \textbf{Avg} \\ \midrule
        \multirow{2}{*}{LLaMA-1 (7B)} & Shared LoRA & 0.052 & 78.0 \\
         & \cellcolor{blue!50} WeGeFT$_{d_{in}}$ & \cellcolor{blue!50}0.052  & \cellcolor{blue!50}\textbf{82.6} \\
        \midrule
        \multirow{2}{*}{Llama-2 (7B)} & Shared LoRA  & 0.052 &  78.3 \\ 
        & \cellcolor{blue!50} WeGeFT$_{d_{in}}$  & \cellcolor{blue!50}0.052 &  \cellcolor{blue!50}\textbf{83.8} \\
        \midrule
         \multirow{2}{*}{Llama-3 (8B)} & Shared LoRA  & 0.044 & 76.1\\ 
        & \cellcolor{blue!50} WeGeFT$_{d_{in}}$ & \cellcolor{blue!50}0.044 & \cellcolor{blue!50}\textbf{87.7} \\
        \bottomrule
    \end{tabular}}
    \label{tab:commonsense-reasoning-shared-lora}\vspace{-2mm}
\end{table}

\section{Remarks on the Effectiveness of WeGeFT}
Based on the experimental results, we may draw intuitive and potentially deeper understanding of PEFT and ReFT methods using pretrained Transformer backbones: Pretrained Transformer backbones ``distill'' general and diverse knowledge from a large-scale pretraining dataset, encoded in the pretrained weights. When fine-tuning them at a downstream task, to ``absorb'' new information in the training data of the downstream task, \textit{one of the simplest updates that minimally ``distorts'' and maximally ``preserves'' the pretrained knowledge} is defined by Eqn.~\ref{eq:waft_in} or Eqn.~\ref{eq:gift_dout}, thanks to the low-rank factorized linear projection in the parameter space. The newly ``absorbed'' information from the downstream task is also linearly expressed in the space spanned by the pretrained weights (knowledge).

\section{Conclusion}
We present Weight-Generative Fine-Tuning (WeGeFT) for adapting pretrained Transformer backbones on downstream tasks. Our WeGeFT learns to generate the fine-tuning weight-residuals for layers selected in fine-tuning directly from their frozen pretrained weights. It is parameterized using two-linear-layers (without bias terms). It achieves multi-faceted efficiency across parameters, representations, compute and memory in comparisons with LoRA and its variants, and ReFT. 
We conduct experiments across various tasks, including Natural Language Generation (instruction following, commonsense reasoning, code generation, and arithmetic reasoning), and Visual Recognition. WeGeFT shows strong performance while retaining multi-faceted efficiency.

\section*{Impact Statement}
This paper presents work whose goal is to advance the field of Machine Learning. There are many potential societal consequences of our work, none of which we feel must be specifically highlighted here.

\section*{Acknowledgments}
This research is partly supported by  NSF IIS-1909644, ARO Grant W911NF1810295, ARO Grant W911NF2210010,  NSF CMMI-2024688 and NSF IUSE-2013451. The views and conclusions contained herein are those of the authors and should not be interpreted as necessarily representing the official policies or endorsements, either expressed or implied, of ARO, NSF or the U.S. Government. The U.S. Government is authorized to reproduce and distribute reprints for Governmental purposes not withstanding any copyright annotation thereon.

\bibliography{example_paper}
\bibliographystyle{icml2025}

\newpage
\appendix
\onecolumn
\section*{Appendix}

\section{Detailed analysis of Training Time}
\label{sec:training-time-analysis}
To show the advantage of WeGeFT over VeRA, we conduct further experiments by setting the intermediate rank in VeRA to be 1024 (as used in \cite{vera}). Table \ref{tab:training-time-analysis} shows that while reducing the dimension lowers the training time and memory costs, it causes a severe drop in accuracy.

\begin{table}[H]
    \centering
    \caption{Results of fine-tuning  Llama 1 and 2 (7B) on the \textbf{Math10k} benchmark. The Mem. refers to GPU memory, and Wall Time is the time required to complete 1 epoch of training. All results are obtained by us using our code base for fair comparisons, except those by DiReFT and LoReFT  using LlaMA 1 are from~\cite{loreft}.}
    \resizebox{0.6\linewidth}{!}{
    \begin{tabular}{l|l|c|c|c|cccc|c}
        \toprule
        & \textbf{Method} & \rotatebox{90}{\textbf{Params (\%)}} & \rotatebox{90}{\textbf{Mem. (GB)}} & \rotatebox{90}{\textbf{Wall Time}} & \rotatebox{90}{\textbf{AQuA}} & \rotatebox{90}{\textbf{GSM8k}} & \rotatebox{90}{\textbf{MAWPS}} & \rotatebox{90}{\textbf{SVAMP}} & \rotatebox{90}{\textbf{Avg. Acc.}} \\ 
        \midrule
        \multirow{9}{*}{\rotatebox{90}{LLaMA-1 (7B)}} & LoRA$^{r=16}$ & 0.416 & 18.01 & 0.43 & 23.5 & 38.5 & 85.3 & 56.4 & \textbf{50.9} \\
        & DoRA$^{r=16}$ & 0.427 & 20.37 & 2.36 & 21.5 & 37.9 & 86.0 & 55.3 & 50.2 \\
        & \cellcolor{blue!50}WeGeFT-Sep$_{d_{in}}$ & \cellcolor{blue!50}0.416 & \cellcolor{blue!50}18.01 & \cellcolor{blue!50}0.46 & \cellcolor{blue!50}23.8 & \cellcolor{blue!50}37.9 & \cellcolor{blue!50}84.5 & \cellcolor{blue!50}54.2 & \cellcolor{blue!50}50.1 \\ \cmidrule{2-10}
        & LoRA$^{r=2}$ & 0.052 & 17.74 & 0.43 & 23.1 & 34.6 & 83.9 & 54.1 & 48.9 \\
        & DoRA$^{r=2}$ & 0.065 & 20.09 & 2.36 & 21.1 & 34.6 & 84.0 & 53.8 & 48.4 \\
        & VeRA$^{r=12288}$ & 0.042 & 20.65 & 9.01 & 21.3 & 34.0 & 82.8 & 50.7 & 47.2 \\
        & VeRA$^{r=1024}$ & 0.015 & 17.80 & 1.15 & 23.0 & 30.5 & 79.1 & 48.4 & 45.2 \\
         & \cellcolor{blue!50} WeGeFT$_{d_{in}}$ & \cellcolor{blue!50}0.052 & \cellcolor{blue!50}17.74 & \cellcolor{blue!50}0.51 & \cellcolor{blue!50}24.3 & \cellcolor{blue!50}36.5 & \cellcolor{blue!50}82.4 & \cellcolor{blue!50}56.9 & \cellcolor{blue!50}\textbf{50.0} \\
        & \cellcolor{blue!50} WeGeFT$_{d_{out}}$ & \cellcolor{blue!50}0.016 & \cellcolor{blue!50}17.71 & \cellcolor{blue!50}0.36 & \cellcolor{blue!50}20.7 & \cellcolor{blue!50}33.0 & \cellcolor{blue!50}80.8 & \cellcolor{blue!50}53.5 & \cellcolor{blue!50}\textbf{47.0} \\
        \midrule
        \multirow{9}{*}{\rotatebox{90}{Llama 2 (7B)}} & LoRA$^{r=16}$ & 0.416 & 18.01 & 0.43 & 24.5 & 43.4 & 86.1 & 57.2 & 52.8 \\
        & DoRA$^{r=16}$ & 0.429 & 20.37 & 2.35 & 24.1 & 41.4 & 87.1 & 57.1 & 52.4 \\
        & \cellcolor{blue!50}WeGeFT-Sep$_{d_{in}}$ & \cellcolor{blue!50}0.416 & \cellcolor{blue!50}18.01 & \cellcolor{blue!50}0.46 & \cellcolor{blue!50}26.1 & \cellcolor{blue!50}42.4 & \cellcolor{blue!50}85.9 & \cellcolor{blue!50}58.6 & \cellcolor{blue!50}\textbf{53.3} \\ \cmidrule{2-10}
        & LoRA$^{r=2}$ & 0.052 & 17.74 & 0.42 & 24.7 & 40.2 & 85.0 & 56.0 & 51.5 \\
        & DoRA$^{r=2}$ & 0.065 & 20.09 & 2.35 & 24.0 & 40.6 & 84.6 & 56.0 & 51.3 \\
        & VeRA$^{r=12288}$ & 0.042 & 20.65 & 9.00 & 23.5 & 38.7 & 85.3 & 54.3 & 50.4 \\
        & VeRA$^{r=1024}$ & 0.015 & 17.80 & 1.15 & 23.6 & 35.5 & 82.1 & 53.3 & 48.6 \\
         & \cellcolor{blue!50}WeGeFT$_{d_{in}}$ & \cellcolor{blue!50}0.052 & \cellcolor{blue!50}17.74 & \cellcolor{blue!50}0.50 & \cellcolor{blue!50}23.6 & \cellcolor{blue!50}42.4 & \cellcolor{blue!50}84.2 & \cellcolor{blue!50}57.4 & \cellcolor{blue!50}\textbf{51.9} \\ 
        & \cellcolor{blue!50}WeGeFT$_{d_{out}}$ & \cellcolor{blue!50}0.016 & \cellcolor{blue!50}17.71 & \cellcolor{blue!50}0.39 & \cellcolor{blue!50}26.1 & \cellcolor{blue!50}38.0 & \cellcolor{blue!50}83.1 & \cellcolor{blue!50}57.3 & \cellcolor{blue!50}\textbf{51.1} \\
        \bottomrule
    \end{tabular}
    }
    \label{tab:training-time-analysis} \vspace{-2mm}
\end{table}

\section{Performance of WeGeFT with mixed precision float16}
\label{sec:float16-analysis}
We conduct additional experiments with LLaMA-1 (7B) using mixed-precision float16 instead of mixed-precision bfloat16. The table below shows that WeGeFT and LoRA experience a similar relative performance drop with float16 while maintaining comparable memory and wall-time, consistent with float16’s known limitations as compared to bfloat16. The performance drop due to float16 can be offset by increasing the number of trainable parameters in WeGeFT. As shown in the table, WeGeFT with a rank of 128 outperforms LoRA even with float16, while using four times fewer parameters. These results further confirm WeGeFT’s compatibility with any device that supports LoRA.

\begin{table}[H]
    \centering
    \caption{Results of fine-tuning  Llama 1 (7B) on the \textbf{Math10k} benchmark with pretrained weights and activations converted to float16 and bfloat16 precisions.}
    \resizebox{0.7\linewidth}{!}{
    \begin{tabular}{l|l|c|c|c|cccc|c}
        \toprule
        & \textbf{Method} & \rotatebox{90}{\textbf{Params (\%)}} & \rotatebox{90}{\textbf{Mem. (GB)}} & \rotatebox{90}{\textbf{Wall Time}} & \rotatebox{90}{\textbf{AQuA}} & \rotatebox{90}{\textbf{GSM8k}} & \rotatebox{90}{\textbf{MAWPS}} & \rotatebox{90}{\textbf{SVAMP}} & \rotatebox{90}{\textbf{Avg. Acc.}} \\ 
        \midrule
        \multirow{2}{*}{bfloat16} & LoRA$^{r=16}$ & 0.416 & 18.01 & 0.43 & 23.5 & 38.5 & 85.3 & 56.4 & \textbf{50.9} \\
         & \cellcolor{blue!50}WeGeFT$_{d_{in}}^{r=64}$ & \cellcolor{blue!50}0.052 & \cellcolor{blue!50}17.74 & \cellcolor{blue!50}0.51 & \cellcolor{blue!50}24.3 & \cellcolor{blue!50}36.5 & \cellcolor{blue!50}82.4 & \cellcolor{blue!50}56.9 & \cellcolor{blue!50}50.0 \\
         \midrule
         \multirow{3}{*}{float16} & LoRA$^{r=16}$ & 0.416 & 18.07 & 0.42 & 21.8 & 37.9 & 84.7 & 57.1 & 50.4 \\
         & \cellcolor{blue!50}WeGeFT$_{d_{in}}^{r=64}$ & \cellcolor{blue!50}0.052 & \cellcolor{blue!50}17.75 & \cellcolor{blue!50}0.43 & \cellcolor{blue!50}22.7 & \cellcolor{blue!50}36.0 & \cellcolor{blue!50}83.5 & \cellcolor{blue!50}54.9 & \cellcolor{blue!50}49.3 \\
         & \cellcolor{blue!50}WeGeFT$_{d_{in}}^{r=128}$ & \cellcolor{blue!50}0.104 & \cellcolor{blue!50}17.80 & \cellcolor{blue!50}0.43 & \cellcolor{blue!50}22.7 & \cellcolor{blue!50}38.5 & \cellcolor{blue!50}84.6 & \cellcolor{blue!50}56.1 & \cellcolor{blue!50}\textbf{50.5} \\
        \bottomrule
    \end{tabular}
    }
    \label{tab:float16-analysis} %
\end{table}

\section{Analysis of WeGeFT vs. LoRA}\label{sec:gradient}

To rigorously distinguish WeGeFT from LoRA, we perform an analytical gradient comparison in the context of fine-tuning Transformer-based architectures.

\subsection{Gradient Analysis}

Consider a simplified Transformer layer with pretrained weights $W^l$ for layer $l$. LoRA fine-tunes by introducing additive low-rank residuals:

\begin{equation}
\hat{W}_{\text{LoRA}}^l = W^l + B^l A^l
\end{equation}
where $B^l$ and $A^l$ are learnable low-rank matrices.

In contrast, consider Eqn.~\ref{eq:waft_in}, WeGeFT fine-tunes through a multiplicative residual explicitly dependent on pretrained weights:

\begin{equation}
\hat{W}_{\text{WeGeFT}}^l = W^l (I + \phi\cdot \psi)
\end{equation}
with shared low-rank matrices $\phi$ and $\psi$.

Let $\mathcal{L}$ denote a scalar loss function (e.g., cross-entropy). For LoRA, the gradient computations with respect to the matrices $A^l$ and $B^l$ are:

\begin{equation}
\frac{\partial \mathcal{L}}{\partial A^l} = (B^l)^\top X’^{l}, \quad \frac{\partial \mathcal{L}}{\partial B^l} = X’^{l}(A^l)^\top
\end{equation}

where $X’^{l} = \left(\frac{\partial \mathcal{L}}{\partial X^l}\right)^\top X^{l-1}$ aggregates local gradient information.

For WeGeFT, gradients for $\phi$ and $\psi$ include information aggregated across layers due to parameter sharing:

\begin{equation}
\frac{\partial \mathcal{L}}{\partial \psi} = \phi^\top \sum_{l} (W^l)^\top X’^{l}, \quad \frac{\partial \mathcal{L}}{\partial \phi} = \sum_{l}(W^l)^\top X’^{l}\psi^\top
\end{equation}

Here, WeGeFT gradients inherently integrate knowledge from pretrained weights across multiple layers, encapsulating broader contextual and structural dependencies than LoRA.

\subsection{Implications for Optimization Dynamics}

\begin{itemize}[leftmargin=*,noitemsep,topsep=0pt]
\item \textbf{Layer-wise vs. Global Updates:} LoRA updates parameters in isolation per layer, restricting interaction. In contrast, WeGeFT updates are global, considering inter-layer correlations and leading to more cohesive and stable optimization trajectories.

\item \textbf{Pretrained Knowledge Utilization:} By explicitly multiplying residuals with pretrained weights, WeGeFT exploits existing model structure, preserving crucial pretrained information, potentially yielding superior convergence and generalization.

\item \textbf{Expressivity and Efficiency Trade-off:} WeGeFT maintains expressivity through multiplicative updates despite substantial parameter sharing, balancing parameter efficiency without compromising learning capacity, unlike traditional additive methods such as LoRA.
\end{itemize}

\subsection{Summary of Advantages}

\begin{itemize}[leftmargin=*,noitemsep,topsep=0pt]
\item \textbf{Improved Parameter Efficiency:} Explicitly leverages pretrained weights to achieve stronger fine-tuning results with fewer learnable parameters.
\item \textbf{Optimized Gradient Flow:} Gradients leverage global information, enabling coordinated fine-tuning across layers.
\end{itemize}

This analysis underpins the empirical advantages of WeGeFT observed in extensive experimentation, highlighting fundamental theoretical distinctions from LoRA.

\section{Implementation Details and Hyperparameter Tuning}
~\label{sec:impl}

In practice, we use a scaling factor of $\frac{\alpha}{r}$ for residuals as done in LoRA~\citep{lora}. We also use dropout \cite{dropout} on the pretrained weights before transforming using WeGeFT parameters as a means of regularization:
\begin{equation}
\hat{W}^l =W^l + \frac{\alpha}{r}\text{Dropout}(W^l) \cdot  \phi\cdot \psi,    \label{eq:gift-scaled}
\end{equation} 
We omit this in the main section for ease of notation and simplicity, as it does not affect the analysis. In experiments, we initialize $\psi$ to all zeros and $\phi$ to Kaiming Uniform initialization \citep{he_init}.

\subsection{Computing Resources and Code}
\label{sec:compute}
All our experiments are run on a single Nvidia A100 GPU. Our code is provided in the supplementary materials.

\subsection{Arithmetic Reasoning}
\label{sec:math-hyperparameters}
On the Math10k, we follow \cite{loreft}, and tune the hyperparameters by fine-tuning the LLaMA-1 (7B) model on the GSM8k dataset \cite{gsm8k} using a separate validation set constructed from the training set, and use the same hyperparamters for Llama-2 (7B). Table \ref{tab:math-hyperparameters} shows the hyperparameters used in our experiments. We perform hyperparameter search using the seed 42, and report the final results by averaging over three runs with seeds 42, 43, and 44. We use a greedy decoding scheme during inference, with a maximum new token number of 512. For experiments on fine-tuning Llama 2 (7B) on MetaMathQA and evaluating on GSM8k, we use the hyperparameters from \cite{lora-ga}, and tune the learning rate on a validation split from Meta-MathQA. We report average scores across 3 runs with seeds 42, 43, 44.

\begin{table}[h!]
    \centering
    \caption{Hyperparameters used for the Math10k experiments. We use greedy sampling following \cite{loreft}}
    \resizebox{0.8\textwidth}{!}{
    \begin{tabular}{ll|l}
        \toprule
        & \textbf{Hyperparameter} & \textbf{Value} \\
        \toprule
        & Max Sequence Length & 512 \\
        & Optimizer & AdamW \\
        & Weight Decay & 0.0 \\
        & LR Scheduler & Linear \\
        & Batch Size & 16 \\
        & Epochs & 3 \\
        \midrule
        \multirow{4}{*}{WeGeFT$_{d_{in}}$} & Learning Rate & $4\times 10^{-4}$ \\
        & Rank & 64 \\
        & Scaling Factor & 128 \\
        & Warmup Ratio & 0.1 \\
        & Dropout & 0.1 \\
        & Fine-Tuned Layers & Query, Key, Value, Up Projection, Down Projection \\
        \midrule
        \multirow{4}{*}{WeGeFT$_{d_{out}}$} & Learning Rate & $7\times 10^{-4}$ \\
        & Rank & 64 \\
        & Scaling Factor & 64 \\
        & Warmup Ratio & 0.06 \\
        & Fine-Tuned Layers & Out Projection, Down Projection \\
        \bottomrule
    \end{tabular}}
    \label{tab:math-hyperparameters}
\end{table}

\begin{table}[h!]
    \centering
    \caption{Hyperparameters used for fine-tuning on  MetaMathQA and evaluating on GSM8k.}
    \resizebox{0.8\textwidth}{!}{
    \begin{tabular}{ll|l}
        \toprule
        & \textbf{Hyperparameter} & \textbf{Value} \\
        \toprule
        & Max Sequence Length & 1024 \\
        & Optimizer & AdamW \\
        & Weight Decay & 0.0 \\
        & LR Scheduler & Cosine \\
        & Batch Size & 32 \\
        & Epochs & 1 \\
        \midrule
        \multirow{4}{*}{WeGeFT} & Learning Rate & $5\times 10^{-4}$ \\
        & Rank & 64 \\
        & Scaling Factor & 128 \\
        & Warmup Ratio & 0.03 \\
        & Dropout & 0.0 \\
        & Fine-Tuned Layers & All linear layers (excluding vocabulary projection and head) \\
        & Generation: Temperature & 0.8 \\
        & Generation: top\_p & 0.95 \\
        \bottomrule
    \end{tabular}}
    \label{tab:metamath-hyperparameters}
\end{table}

\subsection{Commonsense Reasoning}
\label{sec:commonsense-hyperparameters}
We tune the hyperparameters for commonsense reasoning by fine-tuning the LLaMA-1 model on the BoolQ dataset \cite{boolq} using a separate validation set constructed from the training set. Table \ref{tab:commonsense-hyperparameters} shows the hyperparameters used in our experiments. We search for the hyperparameters using LLaMa-1 (7B) and use the same hyperparameters for LLaMA-1 (13B), Llama 2 (7B) and Llama 3 (8B) models. We perform hyperparameter search using the seed 42, and report the final results by averaging over three runs with seeds 42, 43, and 44. We use a greedy decoding scheme during inference, with a maximum new token number of 32.

\begin{table}[h!]
    \centering
    \caption{Hyperparameters used for the commonsense reasoning experiments. We use greedy sampling following \cite{loreft}}
    \resizebox{0.8\textwidth}{!}{
    \begin{tabular}{ll|l}
        \toprule
        & \textbf{Hyperparameter} & \textbf{Value} \\
        \toprule
        & Max Sequence Length & 512 \\
        & Optimizer & AdamW \\
        & Weight Decay & 0.0 \\
        & LR Scheduler & Linear \\
        & Batch Size & 16 \\
        & Epochs & 3 \\
        \midrule
        \multirow{4}{*}{WeGeFT$_{d_{out}}$} & Learning Rate & $9\times 10^{-5}$ \\
        & Rank & 64 \\
        & Scaling Factor & 128 \\
        & Warmup Ratio & 0.1 \\
        & Fine-Tuned Layers & Query, Key, Value, Up Projection, Down Projection \\
        \midrule
        \multirow{4}{*}{WeGeFT$_{d_{out}}$} & Learning Rate & $6\times 10^{-4}$ \\
        & Rank & 64 \\
        & Scaling Factor & 64 \\
        & Warmup Ratio & 0.06 \\
        & Dropout & 0.0 \\
        & Fine-Tuned Layers & Out Projection, Down Projection \\
        \bottomrule
    \end{tabular}}
    \label{tab:commonsense-hyperparameters}
\end{table}

\begin{table}[h!]
    \centering
    \caption{Hyperparameters used for fine-tuning on Code-Feedback and evaluating on HumanEval.}
    \resizebox{0.8\textwidth}{!}{
    \begin{tabular}{ll|l}
        \toprule
        & \textbf{Hyperparameter} & \textbf{Value} \\
        \toprule
        & Max Sequence Length & 1024 \\
        & Optimizer & AdamW \\
        & Weight Decay & 0.0 \\
        & LR Scheduler & Cosine \\
        & Batch Size & 32 \\
        & Epochs & 1 \\
        \midrule
        \multirow{4}{*}{WeGeFT} & Learning Rate & $4\times 10^{-4}$ \\
        & Rank & 64 \\
        & Scaling Factor & 128 \\
        & Warmup Ratio & 0.03 \\
        & Dropout & 0.0 \\
        & Fine-Tuned Layers & All linear layers (excluding vocabulary projection and head) \\
        & Generation: Temperature & 0.8 \\
        & Generation: top\_p & 0.95 \\
        \bottomrule
    \end{tabular}}
    \label{tab:code-hyperparameters}
\end{table}

\subsection{Instruction Following}
\label{sec:instruct-hyperparameters}
For fine-tuning Llama  2 (7B) on WizardLM and evaluating on MT-Bench, we use the hyperparameters from \cite{lora-ga} and use the same learning rate as MetaMathQA experiments. We report average scores across 3 runs with seeds 42, 43, 44.

\subsection{Code Generation}
\label{sec:code-hyperparameters}
For fine-tuning Llama 2 (7B) on Code-Feedback dataset \cite{code-feedback} and evaluating on HumanEval, we use the hyperparameters from \cite{lora-ga} and tune the learning rate on a separate validation split from Code-Feedback. We report average scores across 3 runs with seeds 42, 43, 44.

\begin{table}[h!]
    \centering
    \caption{Hyperparameters used for fine-tuning on WizardLM and evaluating on MT-Bench.}
    \resizebox{0.7\textwidth}{!}{
    \begin{tabular}{l|l}
        \toprule
        \textbf{Hyperparameter} & \textbf{Value} \\
        \toprule
        Max Sequence Length & 1024 \\
        Optimizer & AdamW \\
        Weight Decay & 0.0 \\
        LR Scheduler & Cosine \\
        Batch Size & 32 \\
        Epochs & 1 \\
        \midrule
        Learning Rate & $5\times 10^{-4}$ \\
        Rank & 64 \\
        Scaling Factor & 128 \\
        Warmup Ratio & 0.03 \\
        Dropout & 0.0 \\
        Fine-Tuned Layers & All linear layers (excluding vocabulary projection and head) \\
        Generation: Temperature & 0.8 \\
        Generation: top\_p & 0.95 \\
        \bottomrule
    \end{tabular}}
    \label{tab:wizardlm-hyperparameters}
\end{table}

\subsection{FGVC Experiments}%
\label{sec:impl-vision}
For all the experiments, we use ViT-B/16 model~\cite{vit}, which contains 12 transformer blocks, each with 12 heads in the Multi-Head Self-Attention (MHSA) blocks, and a dimension of 768. We use checkpoints from the model pretrained on the ImageNet21k \cite{imagenet21k} under the supervised training protocol provided by the {\tt timm} package. For both VTAB and FGVC experiments, we use a hyperparameter search using the validation sets and use the training+validation data during the final run and report the results on the test sets. The hyperparameter search space used in our experiments in provided in Table \ref{tab:visual-hyperparameter-space}. We use the same train, validation and test splits as \cite{toast}, \textit{except for} Stanford Cars dataset \cite{cars}. Due to the unavailability of the dataset from the \href{https://ai.stanford.edu/~jkrause/cars/car_dataset.html}{original source}, and the difference in the format of the data provided by the \href{https://www.kaggle.com/datasets/jessicali9530/stanford-cars-dataset}{updated source}, we create our own training and validation split (with the same number of images as \cite{toast}) and use the official testing split. We initialize $\phi$ with zeros and $\psi$ with Kaiming uniform initialization.

\begin{table}[h!]
    \centering
    \caption{Hyperparameter search space used for FGVC experiments. During the search, we use 25 epochs due to computational constraints, and use 100 epochs in the final run with the selected hyperparameters}
    \resizebox{0.65\textwidth}{!}{
    \begin{tabular}{cc|c}
        \toprule
        & \textbf{Hyperparameter} & \textbf{Values} \\
        \toprule
        \multirow{2}{*}{BitFit} & Learning Rate & $1e^{-3}$, $1.5e^{-3}$, $2e^{-3}$, $2.5e^{-3}$, $5e^{-3}$, $1e^{-2}$ \\
        & Weight Decay & $0.0$ \\
        \midrule
        \multirow{2}{*}{VPT} & Learning Rate & $1e^{-3}$, $1.5e^{-3}$, $2e^{-3}$, $2.5e^{-3}$, $5e^{-3}$, $1e^{-2}$ \\
        & Weight Decay & $0.0$ \\
        & Num. Prompts & 5 \\
        \midrule
        \multirow{2}{*}{LoRA} & Learning Rate & $1e^{-3}$, $1.5e^{-3}$, $2e^{-3}$, $2.5e^{-3}$, $5e^{-3}$, $1e^{-2}$ \\
        & Weight Decay & $0.01$, $0.001$, $0.0001$, $0.0$ \\
        & Rank \textit{r} & 8 \\
        \midrule
        \multirow{2}{*}{WeGeFT} & Learning Rate & $1e^{-4}$, $2.5e^{-4}$, $5e^{-4}$, $1e^{-3}$, $2.5e^{-3}$, $5e^{-3}$ \\
        & Weight Decay & $0.01$, $0.001$, $0.0001$, $0.0$ \\
        & Rank \textit{r} & 16 \\
        \midrule
        & Optimizer & AdamW \\
        & LR Scheduler & Cosine \\
        & Warmup Epochs & 5 \\
        & Epochs & 100 \\
        & Batch Size & 32 \\
        \bottomrule
    \end{tabular}}
    \label{tab:visual-hyperparameter-space}
\end{table}

\newpage

\clearpage
\section{Visual Inspection of Our Two-Linear-Layer Parameterized WeGeFT}\label{sec:interpretability} %

\begin{figure}[!h] 
        \centering
   \includegraphics[width=1.\textwidth]{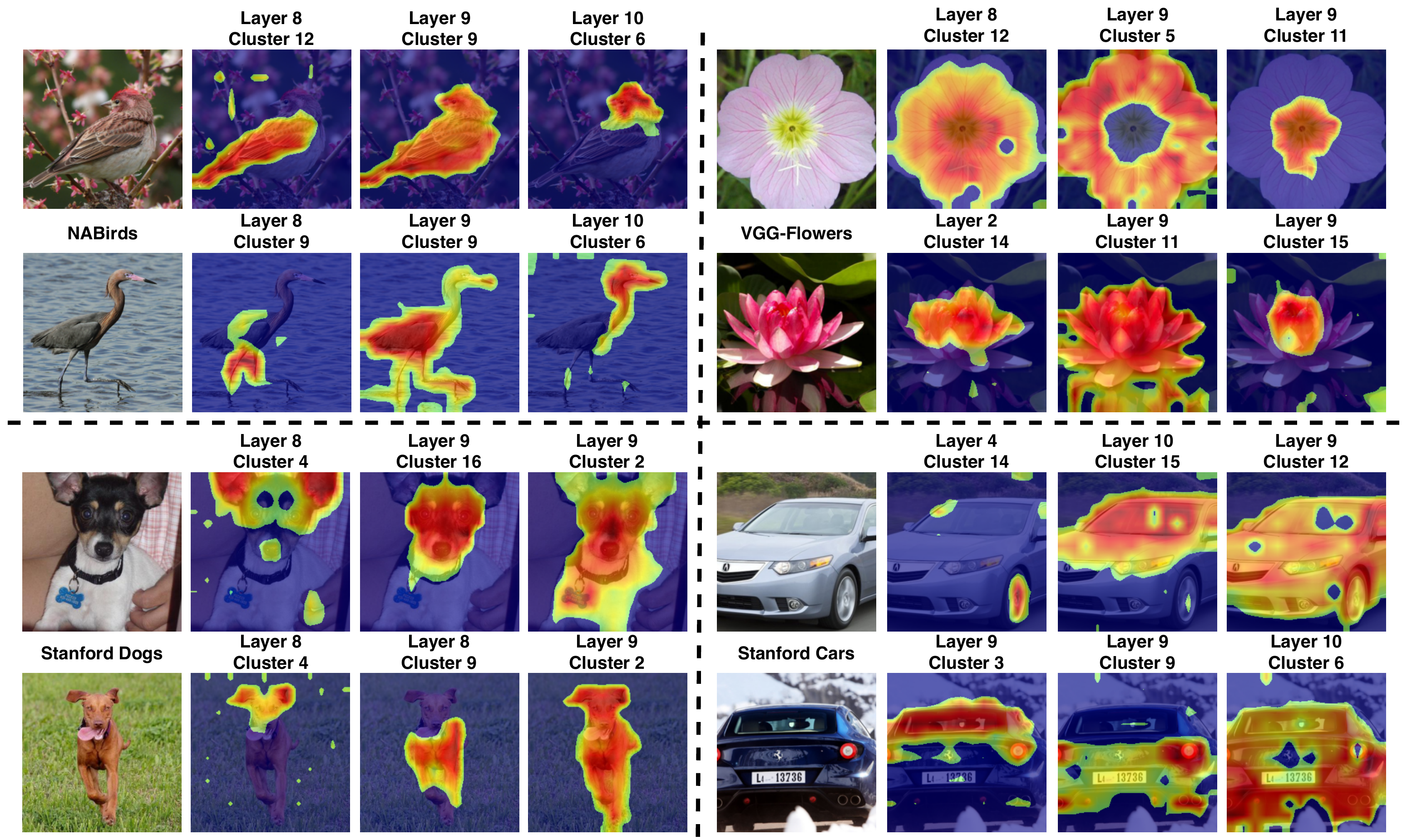} %
   \caption{
   More examples of the visual interpretability of our two-linear-layer parameterized WeGeFT tested on the FGVC benchmark. We show examples of head, wings and legs of birds in the \textit{top-left},  examples of flower petals in the \textit{top-right}, examples of head, ears and legs of dogs in the \textit{bottom-left}, and examples of tires, windshield and bumper of cars in the \textit{bottom-right}. 
   }
   \label{fig:semantic-clusters}\vspace{-1mm}
\end{figure}

\end{document}